\documentclass[conference]{IEEEtran}
\IEEEoverridecommandlockouts

\usepackage{cite}
\usepackage{amsmath,amssymb,amsfonts}
\usepackage{algorithmic}
\usepackage{graphicx}
\usepackage{textcomp}
\usepackage{xcolor}
\usepackage{float}
\usepackage{subcaption}
\usepackage{hyperref}
\usepackage{url}

\def\BibTeX{{\rm B\kern-.05em{\sc i\kern-.025em b}\kern-.08em
    T\kern-.1667em\lower.7ex\hbox{E}\kern-.125emX}}

\begin{document}

\title{Acoustic Anomaly Detection for Early Identification of Drill Jamming%
\thanks{\copyright~2026 IEEE. Accepted for publication in the IEEE International Conference on Industrial Informatics (INDIN 2026).}%
}

\author{
\IEEEauthorblockN{1\textsuperscript{st} Maria Beatriz Silva}
\IEEEauthorblockA{\textit{Federal Rural University of Pernambuco} \\
Recife, Brazil \\
beatriz.asilva@ufrpe.br}
\and
\IEEEauthorblockN{2\textsuperscript{nd} Lucas Edson Silva de Araújo}
\IEEEauthorblockA{\textit{Federal Rural University of Pernambuco} \\
Recife, Brazil \\
lucas.edson@ufrpe.br}
\and
\IEEEauthorblockN{3\textsuperscript{rd} Sergio F. Chevtchenko}
\IEEEauthorblockA{\textit{ICNS, Western Sydney University} \\
Penrith, Australia \\
s.chevtchenko@westernsydney.edu.au}
\and
\IEEEauthorblockN{4\textsuperscript{th} Ermeson Andrade}
\IEEEauthorblockA{\textit{Federal Rural University of Pernambuco} \\
Recife, Brazil \\
ermeson.andrade@ufrpe.br}
\and
\IEEEauthorblockN{5\textsuperscript{th} Danilo Araujo}
\IEEEauthorblockA{\textit{Federal Rural University of Pernambuco} \\
Recife, Brazil \\
danilo.araujo@ufrpe.br}
\and
\IEEEauthorblockN{6\textsuperscript{th} Saeed Afshar}
\IEEEauthorblockA{\textit{ICNS, Western Sydney University} \\
Penrith, Australia \\
s.afshar@westernsydney.edu.au}
}

\maketitle

\begin{abstract}
In the context of Industry 4.0, predictive maintenance is essential to minimize unplanned downtime and ensure operational safety. In machining processes, drill jamming is a critical failure that compromises both workpiece quality and tool integrity. This work proposes an acoustic anomaly detection system using ultrasonic sensors to identify pre-failure patterns in Computer Numerical Control (CNC) machining. We release a public dataset of time- and frequency-domain acoustic features from 17 drills, and benchmark semi-supervised autoencoder-based normality modelling against supervised XGBoost classification using Leave-One-Drill-Out validation. The results demonstrate that the Autoencoder is highly effective for industrial safety, achieving a Recall of 1.00 and a consolidated F1-score of 0.81.  Furthermore, the threshold-independent analysis shows that the Autoencoder better separates normal and pre-failure acoustic patterns, achieving a consolidated AUC of 0.8294, compared with 0.7559 for the supervised model under inter-tool covariate shift. In contrast, the supervised XGBoost achieved an average F1-score of 0.55 and failed to detect failures in some of the evaluated drills. The findings highlight that the ultrasonic spectrum provides a reliable window for identifying incipient mechanical instabilities, and the AE model offers a robust safety envelope by anticipating all jamming events across the tested drill bits.
\end{abstract}

\begin{IEEEkeywords}
Acoustic Emission, Drill Jamming, Anomaly Detection, Autoencoders, XGBoost, Predictive Maintenance.
\end{IEEEkeywords}

\section{Introduction}
In the context of Industry 4.0, predictive maintenance strategies have replaced traditional corrective and preventive approaches by integrating sensors and advanced data analysis techniques into production systems \cite{zaro2022estudo}. In machining operations, tool integrity is a critical factor for product quality and operational safety. Among the most severe events is drill jamming, which can cause damage to the workpiece, the tool, and the machining center, in addition to generating unplanned downtime and high costs \cite{teti2010advanced}.

Acoustic monitoring has emerged as a non-invasive and low-cost alternative for tool condition monitoring. Unlike force or vibration sensors directly coupled to the system, acoustic sensors can be easily integrated into the industrial environment without interfering with the production process. Studies based on acoustic emission show that high-frequency components are particularly sensitive to changes in the cutting regime and to the onset of mechanical instabilities, especially in micro-drilling operations \cite{nakano2023tool}. This sensitivity occurs because acoustic emission signals are generated by the rapid release of strain energy from localized sources such as plastic deformation and crack propagation, which are inherently high-frequency phenomena in machining \cite{teti2010advanced,jemielniak1999commercial}. In this context, acoustic signals in the ultrasonic range present high potential for capturing subtle signatures associated with tool wear and pre-failure phenomena. Therefore, ultrasonic spectrum analysis can reveal patterns that precede catastrophic failures.

Despite advances in tool condition monitoring, relevant challenges remain. In real industrial environments, failures are rare events when compared with the large volume of normal data, resulting in imbalanced datasets that hinder the application of traditional classifiers \cite{chevtchenko2023anomaly, reeber2024tool}. In addition, natural variations in service life and wear dynamics among different tools may introduce changes in the statistical distribution of data between training and testing phases, a phenomenon known as \textit{covariate shift} \cite{sugiyama2007covariate}. This effect compromises the generalization capability of models when applied to tools not seen during training.

Although the literature presents effective methods for fault diagnosis after occurrence, as demonstrated by Nakano et al. \cite{nakano2023tool} and Tagawa et al. \cite{tagawa2021acoustic}, there remains a gap in the investigation of explicit anticipation of jamming under rigorous validation protocols\cite{reeber2024tool}. In particular, many existing studies do not explicitly account for covariate shift across different tools, often relying on validation strategies that mix data from the same tool in both training and testing sets, leading to overestimated performance and limited real-world generalization. To the best of our knowledge, no study simultaneously explores (i) the use of the ultrasonic microphones for drill wear monitoring throughout the tool's life cycle, (ii) inter-tool validation strategies that prevent information leakage between distinct cycles and explicitly simulate covariate shift conditions, and (iii) the systematic comparison between supervised models and anomaly detection approaches with respect to their robustness under distributional shifts.

Given this scenario, this work proposes a modular system for offline detection of ultrasonic acoustic anomalies, composed of independent stages for signal processing, feature extraction, and specialized detection models, aimed at the early identification of drill jamming in CNC machining processes. The study is conducted on a dataset composed of 17 4mm carbide drills, covering different wear profiles and jamming events, whose signals were acquired by an ultrasonic microphone positioned externally to the cutting region. This setup evaluates model robustness to variability across physically distinct tools.
To address the challenges of data imbalance and inter-tool variability, this work evaluates and compares two complementary machine learning paradigms: (i) a semi-supervised model based on an \textit{Autoencoder} trained exclusively with normal data, in which the detection threshold is modeled as a controllable statistical parameter, and (ii) a supervised model based on \textit{XGBoost} with evolutionary hyperparameter optimization. This comparison is particularly relevant under covariate shift conditions, where models must generalize across unseen tool distributions.

To ensure a realistic evaluation of generalization capability, the \textit{Leave-One-Drill-Out}  (LODO) strategy is adopted, equivalent to group-based cross-validation, in which each drill is alternately used as the test set. This configuration prevents data from the same tool from being simultaneously present in training and testing, simulating the industrial scenario in which the model must operate on unseen tools. In this context, LODO explicitly enforces a covariate shift condition between training and testing data.
This dual analysis aims to determine which approach provides a more reliable early warning mechanism, capable of identifying the subtle spectral shifts that precede tool failure. The main contributions of this work are: (i) the construction, documentation, and public release of a multi-drill acoustic dataset, including a detailed description of the experimental protocol; (ii) the proposal of a rigorous inter-tool validation protocol for acoustic monitoring of drills under covariate shift conditions; and (iii) the comparative analysis between supervised and unsupervised approaches, including the investigation of the influence of the decision threshold on \textit{Autoencoder} performance.

\section{Related Work}
\label{rw}
The monitoring of cutting tool condition has been widely investigated under different sensing approaches, including vibration, cutting force, spindle electrical current, and acoustic emission. 
Among these, the analysis of acoustic and ultrasonic signals has proven effective in capturing microscopic phenomena associated with friction, microcracks, and instabilities in the cutting regime.
It is important to distinguish between acoustic emission sensors and microphones; while AE sensors are typically fixed directly to the workpiece or tool (contact coupling) to capture elastic waves, microphones can be installed non-invasively on the machine structure or externally to the cutting region, capturing sound transmitted through the air.

Studies based on acoustic emission show that high-frequency components are particularly informative for progressive tool degradation. Nakano et al. \cite{nakano2023tool} proposed a monitoring method for micro-drilling based on anomaly segmentation in time-frequency images derived from acoustic emission signals, showing that specific spectral signatures emerge in advanced wear stages. Similarly, Tagawa et al. \cite{tagawa2021acoustic} investigated the detection of mechanical faults through acoustic signals in noisy industrial environments, highlighting the need for robust techniques capable of isolating relevant patterns even under significant interference.

In addition to the use of acoustic signals, approaches based on internal machine control data have also been explored. Reeber et al. \cite{reeber2024tool} evaluated anomaly detection methods applied to tool condition monitoring in drilling processes, demonstrating that machine learning models can distinguish normal and anomalous behavior with consistent performance. However, these studies focus predominantly on the identification of already established irregularities, with less emphasis on the explicit anticipation of the jamming event.

In the context of modeling, supervised methods based on ensemble learning, such as \textit{XGBoost}, have shown significant results in industrial fault diagnosis tasks. Wu et al. \cite{wu2021improved} demonstrated that combining \textit{XGBoost} with genetic algorithms for hyperparameter optimization significantly improves performance in electrical fault detection systems. In parallel, semi-supervised and reconstruction-based approaches, such as \textit{Autoencoders}, have been widely employed in scenarios with scarce labeled data \cite{chandola2009anomaly}, modeling normal behavior to identify relevant statistical deviations \cite{sakurada2014}.

Despite these advances, a recurring challenge in industrial applications is the variability between life cycles of different tools. Differences in wear rate and operating conditions may induce changes in the statistical distribution of data between training and testing sets, a phenomenon known as \textit{covariate shift} \cite{sugiyama2007covariate}. This distributional mismatch directly affects the generalization capability of data-driven models when applied to unseen tools.
Many studies report validation by random sampling or traditional \textit{k-fold} procedures \cite{reeber2024tool}, which may introduce information leakage when data from the same tool appear simultaneously in training and testing, overestimating the real performance of the model \cite{sugiyama2007covariate}. As a result, reported performance may not reflect real industrial deployment conditions, where models must operate on previously unseen tools.
Building on these limitations, the present work evaluates Autoencoder and XGBoost architectures to detect imminent drill jamming in rare-failure scenarios under a Leave-One-Drill-Out validation strategy. This approach explicitly enforces inter-tool separation and simulates covariate shift conditions, providing a more realistic assessment of model robustness for industrial deployment.

\section{Methodology}
\label{mt}

This section details the experimental configuration used for data acquisition, acoustic signal processing, and the modeling strategy for early detection of jamming in small-diameter drills.

\subsection{Experimental Setup and Data Acquisition}

The experimental data was collected using a CNC machining center during the drilling of iron plates. Each 4 mm diameter carbide drill followed a programmed sequence of holes until critical failure. However, it is important to emphasize that tool life is inherently stochastic. As discussed by Teti et al. \cite{teti2010advanced}, the transition to tool failure is influenced by multiple factors such as tool-workpiece interaction, material heterogeneity, and process variability, leading to significant uncertainty in tool life.

\begin{figure}[htbp]
    \centering

    \begin{subfigure}{0.6\textwidth}
        \centering
        \includegraphics[width=0.4\textwidth]{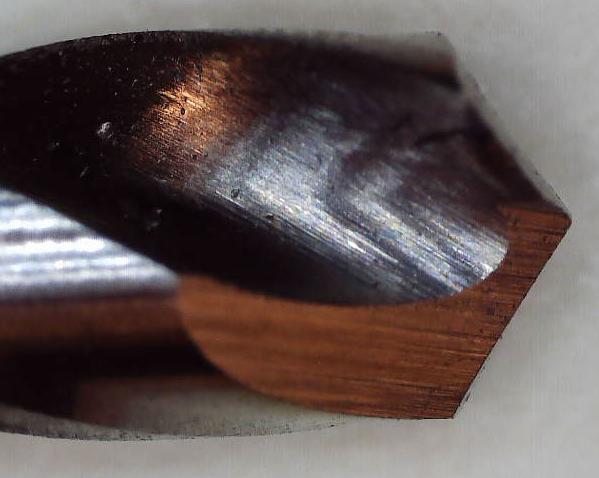}
        \caption{}
        \label{fig:new}
    \end{subfigure}

    \vspace{0.5em}

    \begin{subfigure}{0.6\textwidth}
        \centering
        \includegraphics[width=0.4\textwidth]{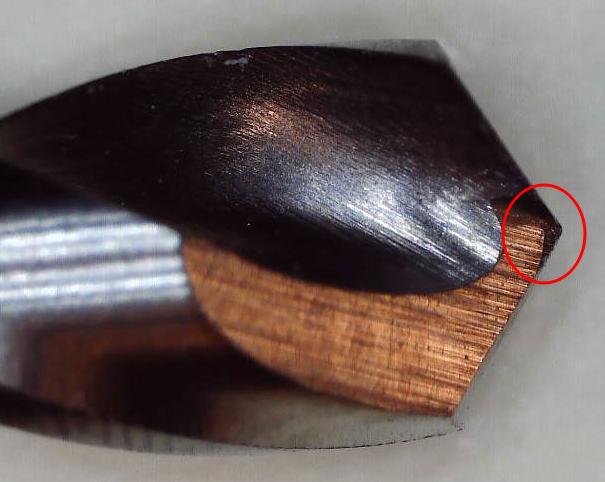}
        \caption{}
        \label{fig:worn_22}
    \end{subfigure}

    \vspace{0.5em}

    \begin{subfigure}{0.6\textwidth}
        \centering
        \includegraphics[width=0.4\textwidth]{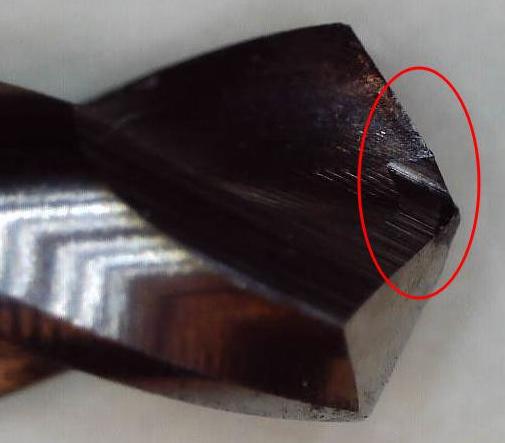}
        \caption{}
        \label{fig:worn_107}
    \end{subfigure}

    \caption{Microscopic inspection showing tool life variability: (a) initial state, (b) premature failure, and (c) advanced wear.}
    \label{fig:drill_comparison}
\end{figure}

This unpredictability is visually demonstrated in Fig. \ref{fig:drill_comparison}. While the new tool (Fig. \ref{fig:drill_comparison}a) presents intact cutting edges, Drill \#10 (Fig. \ref{fig:drill_comparison}b) suffered a jamming event after only 22 holes. In contrast, Drill \#2 (Fig. \ref{fig:drill_comparison}c) sustained 107 holes before reaching a similar state of critical failure. These observations suggest that tool jamming is not strictly correlated with cumulative wear, justifying the need for the real-time anomaly detection approach proposed in this work.

\begin{figure}[H]
    \centering
    \includegraphics[width=0.8\linewidth]{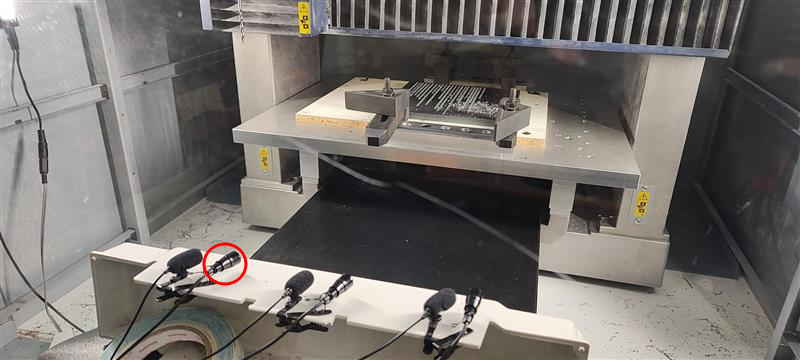}
    \caption{Experimental setup showing the drilling machine, the workpiece material, and the positioning of the acoustic sensors. Three standard-range microphones (96 kHz sampling rate) and three ultrasonic microphones (192 kHz sampling rate) are installed inside the CNC machine enclosure. Data from the highlighted ultrasonic microphone are used in the present study.}
    \label{fig:setup}
\end{figure}

The instrumentation setup focused on the use of ultrasonic Micro-Electro-Mechanical Systems (MEMS) sensors positioned inside the machining area (see Fig. \ref{fig:setup}). This study focuses on internal ultrasonic data because it better captures high-frequency pre-failure signatures, such as friction and micro-cracking, which may be masked by lower-frequency industrial noise in the audible range. Comparisons with microphones in the audible range and fusion of multiple signals are left for future work.

The resulting dataset consists of 17 industrial drills monitored throughout their life cycle. By isolating the ultrasonic signals, the proposed system aims to identify subtle spectral shifts and transient impulses above 20 kHz, which serve as deterministic indicators of the transition from stable operation to the critical wear phase. All scripts and datasets used in this study are publicly available\footnote{\href{\detokenize{https://github.com/BeatrizSilva-Dev/projeto_brocas_AE}}{https://github.com/BeatrizSilva-Dev/projeto\_brocas\_AE}}\footnote{\url{https://doi.org/10.26183/v7kh-b648}}.

\subsection{Dataset and Experimental Configuration}

To ensure statistical robustness and capture operational variability, 17 industrial drills were monitored. All tools were used in the training and evaluation stage of the models, allowing the capture of different degradation patterns and variability among physically distinct tools. Based on the total number of holes drilled until failure, the tools can be categorized into three distinct service life profiles:

\begin{itemize}
\item \textbf{Short Life} ($18 \le n \le 30$ holes): Drills 03, 05, 07, 08, 09, 10, 12, 15, and 16, representing premature failures or accelerated degradation.
\item \textbf{Medium Life} ($31 \le n \le 60$ holes): Drills 01, 04, 06, 11, 13, 14 and 17.
\item \textbf{Long Life} (107 holes): Drill 02, representing a tool with superior operational durability.
\end{itemize}

\subsection{Signal Acquisition and Segmentation}

The acquisition strategy prioritized the capture of high-frequency acoustic signals via an ultrasonic MEMS microphone, mounted inside the CNC machine enclosure, as shown in Figure \ref{fig:setup}. This sensor was selected for its sensitivity to incipient anomalies and transient impulses during the pre-failure stages, enabling the capture of signatures that are absent or attenuated in lower frequency bands.

The recordings were performed continuously and later segmented into individual holes. To ensure high temporal resolution, the signal was processed using a sliding window approach with a frame length of 1024 samples and a hop length of 512 samples, resulting in a 50\% overlap. This configuration provided sufficient temporal resolution to identify the energy valleys associated with drill entry and exit, while preserving transient instabilities critical for anomaly detection.

\subsection{Class Definition}
This work adopted a labeling strategy based on the physical progress of the tool wear and its remaining useful life (RUL), prioritizing an objective criterion over subjective human perception. It was established that the critical wear phase, which precedes the drill jamming, corresponds to the final stage of the tool's operation, resulting in the following definition:

\begin{itemize}
    \item \textbf{Normal Class (0):} Comprises the first 50\% of the drill's lifespan, representing the most stable operating condition. This window is used to define the healthy acoustic baseline.
    
    \item \textbf{Transition Zone (Unlabeled):} Comprises the intermediate 30\% of the tool's lifespan (between the 50\% and 80\% marks). These data points are excluded from the supervised training labels to ensure class purity and avoid training the models with ambiguous degradation patterns.
    
    \item \textbf{Anomalous/Pre-Failure Class (1):} Defined as the final 20\% of holes drilled before tool jamming or breakage, capturing the high-frequency instabilities.
\end{itemize}

Although the Transition Zone is not used for training the supervised models, it remains present during the validation phase to observe the evolution of the anomaly indicators throughout the entire tool life. 
It is also worth noting that while tool degradation is a continuous and stochastic process, using a fixed boundary enables consistent labeling across drills. This categorization ensures that the ground truth reflects the mechanical degradation leading to tool failure, providing a deterministic target for the monitoring models. 

\begin{figure}[H]
    \centering
    \includegraphics[width=\linewidth]{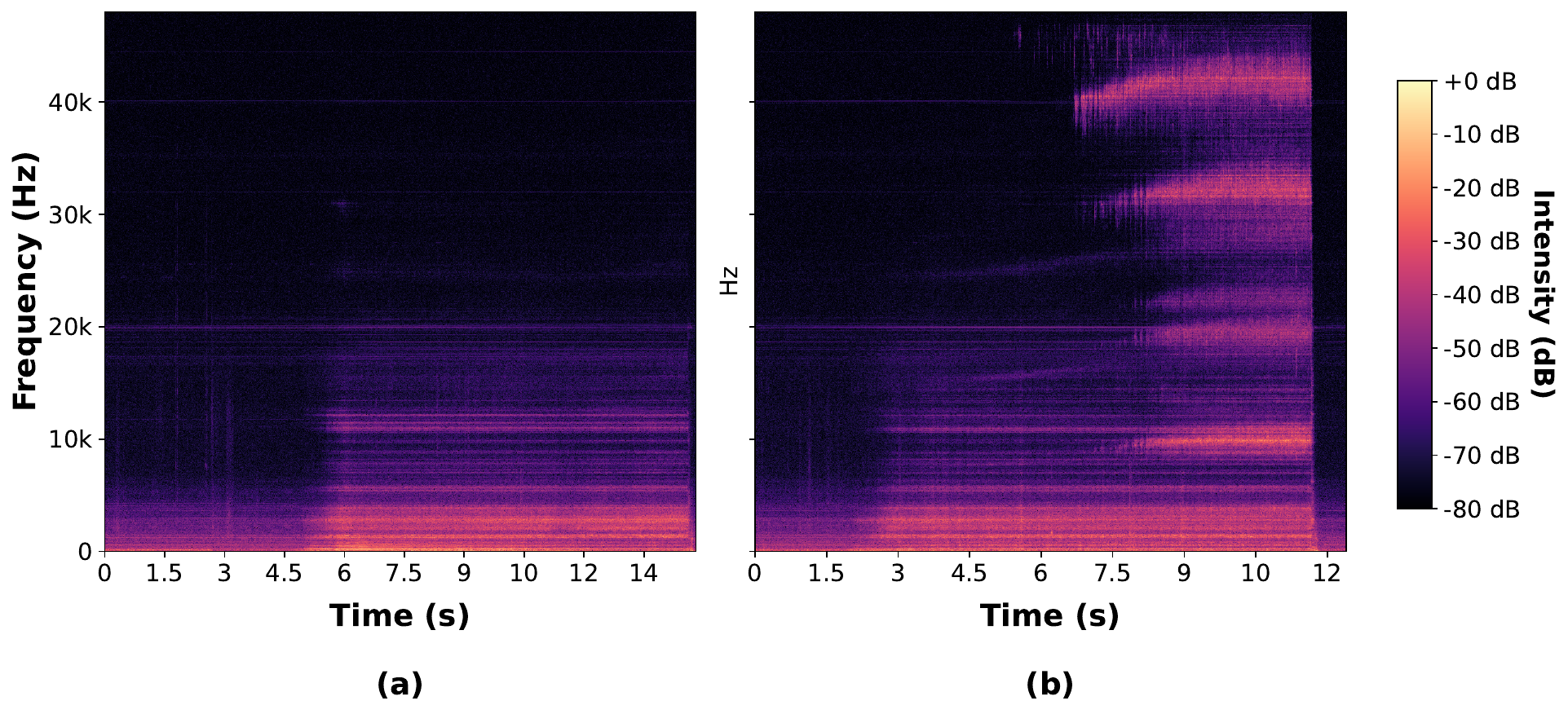}
    \caption{Spectral comparison between (a) a stable drilling condition (first hole) and (b) hole \#24 within the critical wear zone (Last 20\% of tool life).}
    \label{fig:spectrogram_comparison}
\end{figure}

As illustrated in Figure~\ref{fig:spectrogram_comparison}, the normal condition (a) presents a stable energy distribution. In contrast, the critical wear phase represented by hole \#24 (b) is characterized by intense high-frequency transient impulses. These bursts of energy, predominantly occurring above 20 kHz, indicate the severe friction and mechanical instability that precede the jamming event.

\subsection{Feature Extraction}
\label{ssec_fe}
To represent the tool's acoustic signature, signals captured by the MEMS ultrasonic sensor, highly sensitive to micro-cracks and friction, were decomposed into a 46-dimensional feature vector. This multi-domain representation combines descriptors in both time and frequency domains to robustly capture the energy intensity and spectral shifts associated with tool degradation \cite{Chang2025,yuan2023material}. The feature vector was constructed by calculating the mean and standard deviation for each of the following 23 metrics over the analyzed segments \cite{arslan2016tool}:

\begin{itemize}
\item \textbf{MFCCs (Mel-Frequency Cepstral Coefficients, 20 coefficients):} Employed to prioritize critical energy bands through a logarithmic Mel scale based on auditory perception, resulting in 40 features;
\item \textbf{RMS Energy:} Used to quantify variations in signal amplitude and average power, representing the intensity of the acoustic signal (2 features);
\item \textbf{Spectral Centroid:} Related to the center of gravity of the energy distribution across frequencies (2 features);
\item \textbf{Spectral Bandwidth:} Indicating the spectral dispersion and the range of frequencies present in the signal (2 features).
\end{itemize}

This 46-dimension descriptor ($2 \times 23$) ensures that subtle signatures of mechanical instability, predominantly occurring in the ultrasonic range above 20 kHz, are effectively captured for the subsequent anomaly detection process.

\subsection{Machine Learning Architectures}

In this work, two complementary machine learning paradigms are evaluated and compared: a semi-supervised Autoencoder for novelty detection and a supervised XGBoost classifier. This dual analysis aims to determine which approach, normality modeling or boundary-based supervised learning, provides more reliable detection of the spectral changes associated with tool degradation.

\subsubsection{Semi-Supervised Approach (Autoencoder with Dynamic Threshold)}
An \textit{Autoencoder} is a neural network trained to learn a compact representation of input data. In this work, the model was trained exclusively with data from the stable operating regime.

\begin{itemize}
    \item \textbf{Operation:} The model is trained to reconstruct the 46-dimensional feature vector detailed in Section \ref{ssec_fe}. During testing, holes that present a reconstruction error (MSE) significantly higher than the learned healthy pattern are classified as anomalies based on the thresholding criteria described below.
    \item \textbf{Architecture:} MLP Regressor with hidden layers of (32, 16, 32) neurons and ReLU activation.
    \item \textbf{Training Set Definition:} Following the LODO strategy, for each test tool, the model is trained using only the first $N_{normal} = 5$ initial holes of 16 drills at a time, while keeping the last one for testing. This approach ensures the characterization of a general healthy state, avoiding contamination by wear signals.
    \item \textbf{Evaluation Logic:} To evaluate performance, all holes of the unseen test drill are analyzed. For metric calculation, the performance is assessed by comparing the model's output against the defined Normal (first 50\%) and Anomalous (last 20\%) regions, ensuring a clear distinction between tool states. The model's success is determined by its ability to generate an MSE that exceeds the statistical threshold during the transition to the final phase.
    \item \textbf{Threshold Calibration:} To ensure robust generalization and avoid data leakage, a statistical threshold is defined based on the reconstruction error distribution of the training set. We utilized the 99.5th percentile of the training MSE to establish a conservative limit that minimizes false alarms while maintaining maximum sensitivity to incipient failures.
\end{itemize}

\subsubsection{Supervised Approach (\textit{XGBoost})}

For the supervised approach, the \textit{Extreme Gradient Boosting} (\textit{XGBoost}) algorithm was used, a learning method based on decision-tree gradient boosting. In this model, a sequential set of trees is constructed, in which each new tree is trained to correct the errors of the previous ones. This strategy makes it possible to capture complex nonlinear relationships between the acoustic features extracted from the signal and the tool state.

The model receives the same 46-dimensional descriptor detailed in Section \ref{ssec_fe} as input and produces the probability of an anomaly associated with the critical wear phase. To ensure consistency with the established labeling logic, the model was trained to distinguish between the stable operating region (first 50\% of life) and the critical wear phase (final 20\% of life). To handle class imbalance, the \textit{scale\_pos\_weight} parameter was calculated dynamically, considering the ratio between normal and anomalous samples. A standard decision threshold of 0.5 was adopted for final classification.

\subsection{Evolutionary Hyperparameter Optimization of \textit{XGBoost}}
To ensure a fair comparison with the AE and maximize the performance of the \textit{XGBoost} classifier, stochastic optimization was employed using the DEAP library \cite{fortin2012deap}. Genetic Algorithms (GA) were chosen due to the nonlinear and multidimensional nature of the hyperparameter search space, where exhaustive methods like \textit{Grid Search} would be computationally inefficient. A single-objective approach was adopted to maximize the \textit{F1-Score} on a validation set composed of the established regions: the stable operating phase (first 50\%) and the critical wear phase (final 20\%). The configuration of the genetic operators followed the definitions detailed below:

\begin{itemize}
\item \textbf{Population:} 20 individuals.
\item \textbf{Generations:} 10 (fixed stopping criterion).
\item \textbf{Selection:} tournament selection with size $k=3$, chosen for its ability to maintain selection pressure and robustness against noise in the fitness function \cite{miller1995};
\item \textbf{Crossover:} \textit{Blend Crossover} ($\alpha = 0.2$).
\item \textbf{Mutation:} Gaussian ($\mu = 0$, $\sigma = 0.1$, $P_m = 0.2$).
\end{itemize}

The fitness function employs cross-validation with 5 partitions, using \textit{GroupKFold} to ensure that hyperparameter selection is robust to the variability between different drills. Once the optimal parameters were determined, they were fixed for the final performance evaluation using the LODO strategy. 

The hyperparameter search space is defined as shown in Table \ref{tab:search_space}. The parameters \textit{N. Estimators} and \textit{Max Depth} control the model complexity, balancing learning capacity and generalization. Furthermore, adjusting \textit{Subsample} and \textit{Colsample Bytree} between 0.6 and 1.0 introduces controlled stochasticity, making the classifier more resilient to transient instabilities and background noise inherent to the CNC machining environment.

\begin{table}[H]
\caption{Hyperparameter Search Space (\textit{XGBoost})}
\begin{center}
\begin{tabular}{|l|c|c|}
\hline
\textbf{Hyperparameter} & \textbf{Search Interval} & \textbf{Scale Type} \\
\hline
N. Estimators & [50, 450] & Linear \\
\textit{Max Depth} & [3, 15] & Linear \\
\textit{Learning Rate} & [0.001, 0.1] & Logarithmic \\
\textit{Subsample} & [0.6, 1.0] & Linear \\
\textit{Colsample Bytree} & [0.6, 1.0] & Linear \\
\hline
\end{tabular}
\label{tab:search_space}
\end{center}
\end{table}

\subsection{Evaluation Metrics}
Because class imbalance makes accuracy unreliable, the \textit{F1-score} was adopted as the primary evaluation metric:

\begin{equation}
F1 = 2 \cdot \frac{\text{Precision} \cdot \text{Recall}}{\text{Precision} + \text{Recall}}
\end{equation}

This metric penalizes both false negatives (missed failures) and false positives (false alarms), and is complemented by analysis of the confusion matrix, calculated strictly over the defined Normal (50\%) and Anomalous (20\%) regions.

To evaluate the models’ discriminative capacity independently of a fixed decision threshold, the Receiver Operating Characteristic (ROC) curve and its Area Under the Curve (AUC) were analyzed \cite{hanley1982meaning, fawcett2006introduction}. The ROC curve describes the trade-off between the true positive rate and the false positive rate across different classification thresholds. The AUC summarizes this curve into a single threshold-independent measure, representing the probability that a randomly selected pre-failure sample receives a higher anomaly score than a randomly selected normal sample. This analysis complements the fixed-threshold metrics by assessing whether the models preserve class separation beyond a specific calibration choice.

\section{Results and Discussion}
\label{res}

The experiments were conducted under the LODO strategy, allowing evaluation of the actual generalization capability of the system in the presence of natural variability in tool life. In this configuration, each iteration uses one complete drill as the test set while the remaining drills are used for training, simulating a real industrial scenario in which the model must operate on unseen tools whose wear patterns have not been previously observed.

\subsection{Global Performance Analysis}

The global performance evaluation, conducted under the LODO strategy, demonstrates that both methods effectively capture the ultrasonic signatures associated with tool wear. Following the feature extraction and labeling criteria established in Section \ref{mt} , the system achieved consolidated average \textit{F1-Scores} of 0.81 for the \textit{Autoencoder} and 0.55 for \textit{XGBoost}.

These results highlight the superior sensitivity of the semi-supervised approach. The \textit{Autoencoder}, by strictly modeling the stable operating regime (first 50\% of life), proved highly effective at identifying the onset of mechanical instability. In contrast, the supervised \textit{XGBoost} showed lower performance, due to the difficulty of learning a fixed decision boundary across different tools in a high-dimensional feature space. Furthermore, the \textit{Autoencoder} achieved a global Recall of 1.00, ensuring that all tool jamming events were preceded by an anomaly alert, which is critical for preventing damage to the CNC machinery and workpieces.

To complement these fixed-threshold metrics and assess the model's discriminative power independently of a single decision boundary, a consolidated AUC analysis was conducted across the 17 drills. Within the monitored ultrasonic spectrum, the semi-supervised Autoencoder achieved a consolidated AUC of 0.8294, significantly outperforming the supervised XGBoost classifier, which achieved an AUC of 0.7559. These fixed-threshold and threshold-independent results suggest that, in this experimental setting, modeling the normal acoustic baseline was more effective for detecting pre-failure behavior under inter-tool variability.

\begin{table}[htbp]
\caption{Consolidated Performance Metrics Under Ultrasonic Sensing}
\label{tab:global_metrics}
\centering
\begin{tabular}{lccc}
\hline
\textbf{Model Paradigm} & \textbf{F1-Score} & \textbf{Global Recall} & \textbf{Consolidated AUC} \\ \hline
\textbf{MLP Autoencoder} & 0.81               & 1.00                   & 0.8294                    \\ 
\textbf{XGBoost Classifier} & 0.55            & 0.47                   & 0.7559                    \\ \hline
\end{tabular}
\end{table}

\subsection{Autoencoder Performance}

\begin{figure}[htbp]
    \centering
    \begin{subfigure}[b]{0.495\columnwidth}
        \centering
        \includegraphics[width=1.05\textwidth]{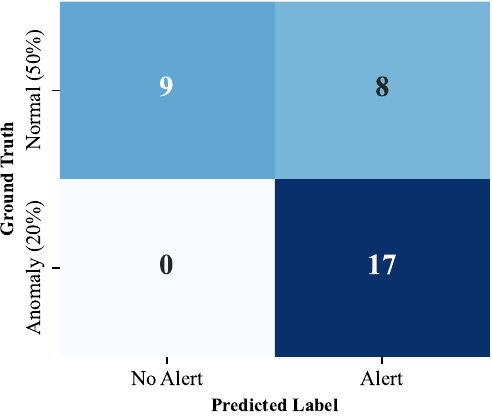} 
        \caption{MLP Autoencoder}
        \label{fig:cm_ae}
    \end{subfigure}
    \hspace{-2mm} 
    \begin{subfigure}[b]{0.495\columnwidth}
        \centering
        \includegraphics[width=1.05\textwidth]{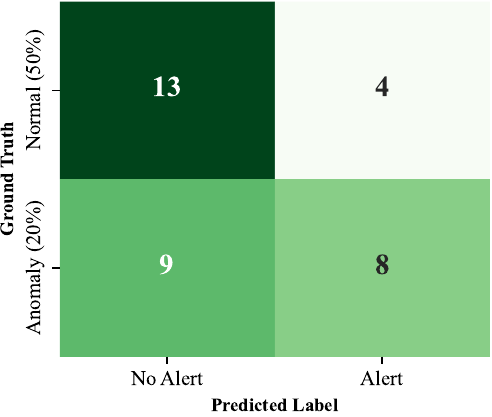} 
        \caption{XGBoost Classifier}
        \label{fig:cm_xgb}
    \end{subfigure}
    \caption{Consolidated confusion matrices across all test drills evaluated under the strict Leave-One-Drill-Out (LODO) validation paradigm.}
    \label{fig:consolidated_confusion_matrices}
\end{figure}

Figure~\ref{fig:cm_ae} presents the consolidated confusion matrix for the Autoencoder across the 17 analyzed drills. The results demonstrate the model's high reliability for industrial safety. Of the 17 drills that reached the critical wear stage (Anomaly Region), all 17 were correctly identified (True Positives), resulting in a Recall of 1.00. The matrix also shows that for the 17 drills in their stable phase (Normal Region), the model correctly classified 9 as ``No Alert'' (True Negatives) and generated 8 early warnings (False Positives). It should be noted that the statistical threshold (99.5th percentile) was deliberately calibrated to be conservative, prioritizing operational safety. This trade-off is often acceptable in predictive maintenance, where an early warning prompts a risk-free preventive verification rather than a machine breakdown.

\subsection{\textit{XGBoost} Model Performance}

For comparison purposes, a supervised model was evaluated using the XGBoost algorithm. Unlike the \textit{Autoencoder}, which learns only the healthy operating regime, \textit{XGBoost} utilizes explicit labels to attempt to learn a decision boundary between the stable phase and the final critical wear phase.

Figure~\ref{fig:cm_xgb} presents the consolidated confusion matrix for the \textit{XGBoost} model across the 17 drills. The results show that \textit{XGBoost} correctly identified 13 normal cases (True Negatives) and only 8 anomalous cases (True Positives). Most critically, the model failed to detect 9 anomalous events (False Negatives), resulting in a significantly lower Recall compared to the semi-supervised approach.

From the perspective of industrial predictive maintenance, these results highlight a significant risk. The \textit{Autoencoder's} null false negative rate is particularly significant for industrial maintenance. While \textit{XGBoost} attempts to establish a precise classification boundary, its failure to generalize across different tools (Recall = 0.47) suggests that tool degradation is more accurately captured as a deviation from a known healthy state than as a fixed discrete class. This confirms that the semi-supervised approach provides a more reliable safety envelope for CNC operations.

\subsection{Inter-Tool Generalization}

\begin{figure}[htbp]
\centering
\includegraphics[width=0.8\linewidth]{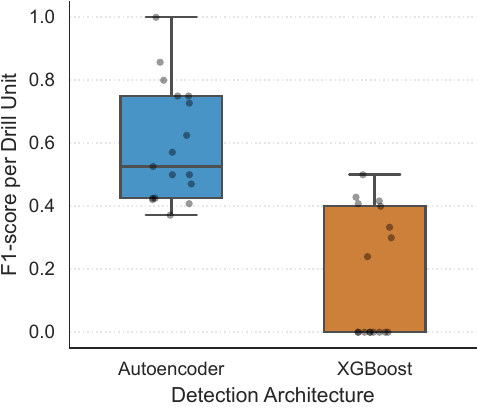} 
\caption{\textit{F1-score} consistency per model across all test drills.}
\label{fig:f1_por_broca}
\end{figure}

Figure~\ref{fig:f1_por_broca} presents the performance distribution across the 17 drills using a boxplot. Under the LODO strategy, this visualization highlights the robust generalization capability of the \textit{Autoencoder} compared to the supervised approach. The Autoencoder exhibits a higher and more consistent performance, with several tools achieving an \textit{F1-score} near 1.0, and a median significantly above the \textit{XGBoost} results.

In contrast, the \textit{XGBoost} model presents a significantly higher variance and a lower median. A critical observation is the presence of several tools with an \textit{F1-score} of zero in the \textit{XGBoost} distribution, corresponding to the false negatives previously discussed. This instability suggests that the supervised approach struggles to establish a universal decision boundary that accounts for the individual wear dynamics of each drill.

The analysis confirms that the \textit{Autoencoder} maintains superior performance regardless of tool durability. The statistical significance of this superiority was verified through a Wilcoxon signed-rank test, which yielded a p-value of $p < 0.001$, confirming that the semi-supervised approach provides a consistently more reliable detection envelope than the supervised model. Even tools with distinct profiles exhibit stable detection patterns, reinforcing the potential of ultrasonic monitoring without the need for tool-specific calibration.

\subsection{Temporal Evolution Analysis}

\begin{figure}[htbp]
    \centering
    \begin{subfigure}[b]{0.9\linewidth}
        \centering
        \includegraphics[width=\linewidth]{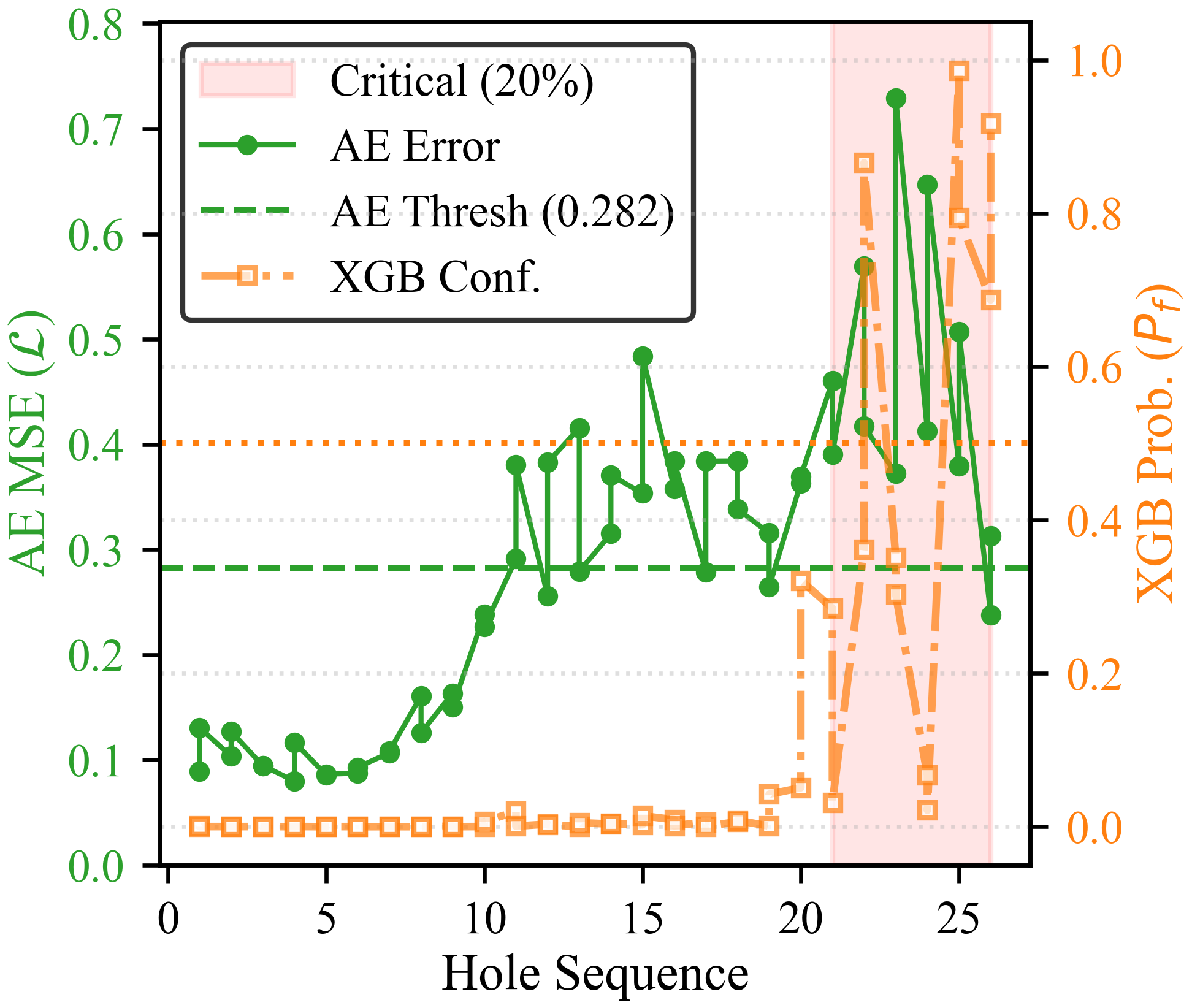}
        \caption{Drill \#03 (Short Life - 26 holes).}
        \label{fig:evolucao_drill03}
    \end{subfigure}
    
    \vspace{10pt}
    
    \begin{subfigure}[b]{0.9\linewidth}
        \centering
        \includegraphics[width=\linewidth]{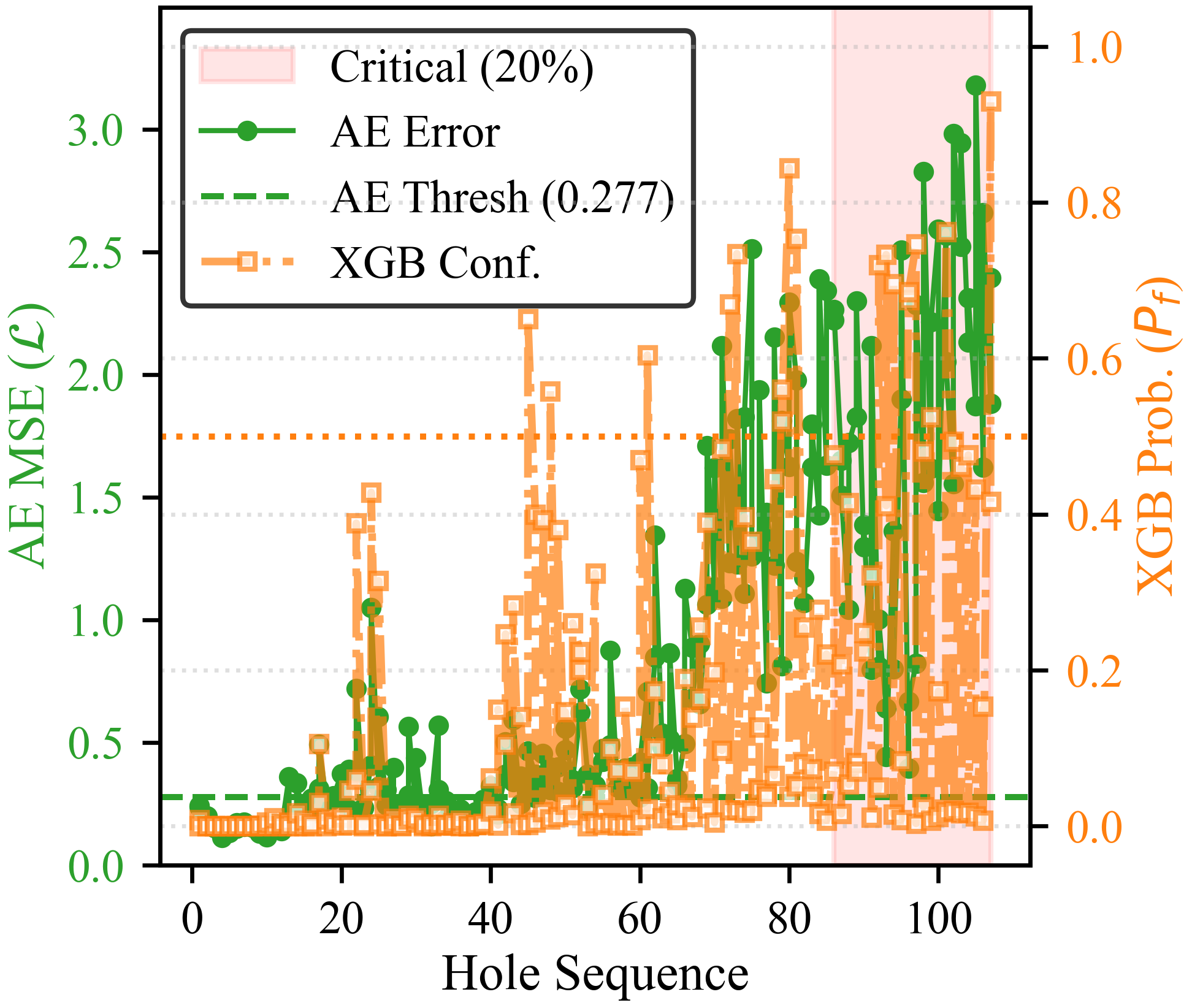}
        \caption{Drill \#02 (Long Life - 107 holes).}
        \label{fig:evolucao_drill05}
    \end{subfigure}
    
    \caption{Temporal evolution of anomaly indicators for unseen tools with different life cycles. The shaded red area represents the critical pre-failure zone (final 20\% of life).}
    \label{fig:temporal_comparison}
\end{figure}

Figure~\ref{fig:temporal_comparison} illustrates the detection dynamics for two unseen drills with distinct service life profiles under the LODO strategy: (a) Drill \#03 (Short Life) and (b) Drill \#02 (Long Life). The left y-axis represents the Autoencoder reconstruction error (MSE), while the right y-axis shows the XGBoost anomaly probability. The dashed horizontal lines indicate the decision thresholds: a statistically defined AE threshold (0.282 in (a) and 0.277 in (b)) and a fixed XGBoost classification threshold of 0.5.

\begin{itemize}
    \item \textbf{Autoencoder Response (Green Line):} 
    In Fig.~\ref{fig:temporal_comparison}(a), the AE reconstruction error remains low and stable during the initial holes, followed by a gradual increase as the tool approaches failure. The threshold crossing occurs close to the critical region, indicating sensitivity to degradation even in short life cycles. 
    In contrast, Fig.~\ref{fig:temporal_comparison}(b) shows a clearer temporal pattern: after approximately hole 60, the AE establishes a persistent alert regime, maintaining values consistently above the threshold well before the final 20\% of tool life.

    \item \textbf{XGBoost Response (Orange Line):} 
    In Fig.~\ref{fig:temporal_comparison}(a), the XGBoost confidence remains close to zero for most of the tool life and only increases abruptly near the critical region, with strong oscillations around the 0.5 threshold. 
    In Fig.~\ref{fig:temporal_comparison}(b), this instability becomes even more evident, with frequent fluctuations above and below the decision boundary, indicating low temporal consistency.

    \item \textbf{Sensitivity Analysis:} 
    In both cases (a) and (b), the AE threshold is crossed before the red shaded region (final 20\%), indicating early detection of pre-failure behavior. These detections correspond to incipient mechanical instabilities captured in the ultrasonic spectrum, rather than false alarms, providing a useful anticipation window for predictive maintenance.
\end{itemize}
    
\subsection{Discussion and Limitations}
The experimental results confirm that high-frequency acoustic monitoring extends beyond the limits of human perception in machining environments. The superior performance of the semi-supervised \textit{Autoencoder} ($F1=0.81$) over the supervised \textit{XGBoost} ($F1=0.55$) represents a significant contribution to the field of tool condition monitoring. It suggests that tool degradation is more effectively characterized as a deviation from a known healthy state rather than a distinct, repeatable class. This is particularly relevant in the ultrasonic domain, where micro-fractures and friction create non-linear spectral shifts that vary significantly between different physical tools.

The main contribution of this work is the adaptation of anomaly detection models to a rigorous inter-tool validation framework. Unlike random data splitting, the LODO strategy evaluates whether the model can generalize to unseen tools. The zero false-negative rate achieved by the \textit{Autoencoder} indicates a reliable safety envelope, which is particularly important in CNC drilling, where missing an impending failure can be far more costly than raising a false alarm. 

Tool degradation is a continuous physical process, while the labeling strategy uses discrete thresholds as a necessary simplification. Furthermore, the thresholds used in this study are specific to 4 mm carbide drills, and applying the approach to other tool geometries or cutting conditions would likely require adaptive calibration to account for different acoustic baselines.

Although the dataset includes only 17 drills, each tool was monitored continuously throughout its operational life until failure. This produced a large dataset with segmented acoustic windows capturing the progression from stable operation to pre-failure. As a result, the dataset provides meaningful operational coverage for evaluating inter-tool generalization, despite the limited number of tools.

\section{Conclusion}
\label{con}

This work demonstrated the value of high-frequency acoustic monitoring for early drill-jamming detection. Under LODO validation, the semi-supervised \textit{Autoencoder} achieved perfect Recall (1.00), an \textit{F1-Score} of 0.81, and significantly outperformed \textit{XGBoost} ($p < 0.001$). Its stronger threshold-independent performance, with an AUC of 0.8294 versus 0.7559 for \textit{XGBoost}, further shows that learning the normal acoustic baseline is more robust under inter-tool variability. These results confirm that ultrasonic signals capture early pre-failure signatures and can provide a reliable anticipation window for predictive maintenance while avoiding tool-specific calibration.

As future work, we suggest the investigation of adaptive thresholding techniques to further refine model sensitivity under varying cutting parameters. Additionally, exploring deep learning architectures for time-series forecasting, such as long short-term memory networks or transformer models, could provide more precise estimates of the RUL, further enhancing decision-making in autonomous Industry 4.0 environments.

\section*{Acknowledgment}
This work was supported by the Conselho Nacional de Desenvolvimento Científico e Tecnológico (CNPq), Brazil, under Grant No. 401147/2025-8. The authors also acknowledge CNPq for the financial support provided through a student scholarship.

\bibliographystyle{IEEEtran}
\bibliography{references}

\end{document}